\documentclass[10pt,twocolumn,letterpaper]{article}

\usepackage[pagenumbers]{cvpr} 

\usepackage{graphicx}
\usepackage{amsmath}
\usepackage{amssymb}
\usepackage{booktabs}
\usepackage{float}
\usepackage[accsupp]{axessibility} 
\usepackage[title]{appendix}
\usepackage[pagebackref,breaklinks,colorlinks]{hyperref}
\usepackage{marvosym}

\usepackage[capitalize]{cleveref}
\crefname{section}{Sec.}{Secs.}
\Crefname{section}{Section}{Sections}
\Crefname{table}{Table}{Tables}
\crefname{table}{Tab.}{Tabs.}

\newcommand{\sysname}{\textbf{ReEF}}
\newcommand{\sysnameb}{\textbf{ReEF} }

\def\confName{CVPR}
\def\confYear{2022}

\begin{document}

\title{Registering Explicit to Implicit: Towards High-Fidelity Garment mesh Reconstruction from Single Images}

\author{Heming Zhu\textsuperscript{1,2} \quad Lingteng  Qiu\textsuperscript{1,2} \quad  Yuda Qiu\textsuperscript{1} \quad Xiaoguang Han\textsuperscript{1,3 \Letter} \\
\textsuperscript{1} SSE, CUHKSZ\quad
\textsuperscript{2} SRIBD\quad
\textsuperscript{3} FNii, CUHKSZ\\
{\tt\small hanxiaoguang@cuhk.edu.cn}
}
\twocolumn[{%
\renewcommand\twocolumn[1][]{#1}%
\maketitle
\begin{center}
    \centering
    \vspace{-2em}
    \includegraphics[width=0.99\textwidth]{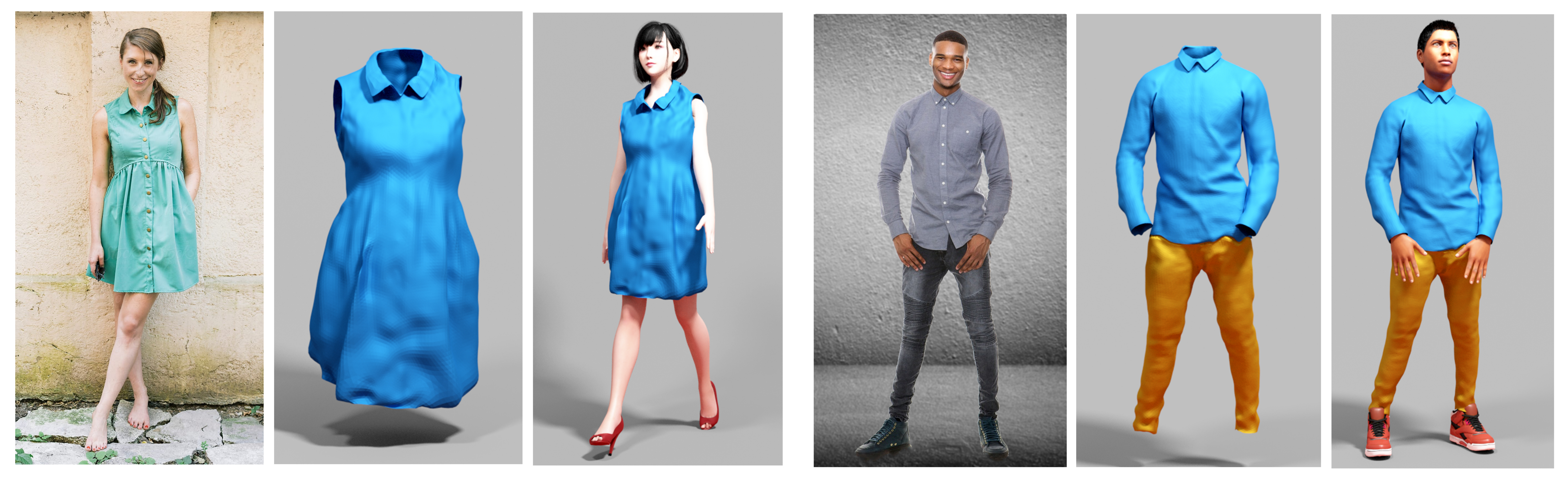}
    \captionof{figure}{Given a single in-the-wild clothed human image, \sysnameb can generate high-fidelity layered garment meshes. The appearances of the reconstructed garments are well aligned with the input image. Moreover, the produced garments can be placed on other virtual characters.
    }
\end{center}%
}]
\vspace{-4mm}
\begin{abstract}
Fueled by the power of deep learning techniques and implicit shape learning, recent advances in single-image human digitalization have reached unprecedented accuracy and could recover fine-grained surface details such as garment wrinkles. However, a common problem for the implicit-based methods is that they cannot produce separated and topology-consistent mesh for each garment piece, which is crucial for the current 3D content creation pipeline. To address this issue, we proposed a novel geometry inference framework \sysnameb that reconstructs topology-consistent layered garment mesh by \textbf{re}gistering the \textbf{e}xplicit garment template to the whole-body implicit \textbf{f}ields predicted from single images. Experiments demonstrate that our method notably outperforms the counterparts on single-image layered garment reconstruction and could bring high-quality digital assets for further content creation.
\end{abstract}
\vspace{-4mm}

\section{Introduction}
\label{sec:intro}

High-quality human-related 3D contents are highly demanded by various real-world applications, including virtual live-streaming, gaming, and filming. However, producing visually plausible 3D digital human assets has always been a laborious task, which may take hours even for an expert modeler. 

In contrast, in-the-wild images are easily accessible with commercial cameras and from the internet. Therefore, recent researches have extensively studied human digitalization from single in-the-wild images, aiming at assisting people without expertise to generate visually plausible 3D human-related contents efficiently. 

Compared with the recent advances in single image body ~\cite{anguelov2005scape,loper2015smpl,hmrKanazawa17,hasler2009statistical,pons2015dyna,lassner2017unite,omran2018neural,rhodin2016general,pare,spec} and clothed human reconstruction~\cite{natsume2019siclope,varol2018bodynet,saito2019pifu,alldieck2019tex2shape,lazova3dv2019,alldieck19cvpr,pumarola20193dpeople,chen2013deformable,tang2019neural}, research on single image layered garment reconstruction is quite sparse. 
The main challenges towards a high-fidelity garment reconstruction are as two folds: \textbf{generating garment styles} and \textbf{recovering surface details}. 
To generate garments with different styles, Multi-Garment Net(MGN)~\cite{bhatnagar2019multi}, BCNet~\cite{jiang2020bcnet} and SMPLicit~\cite{smplicit} adopted either explicit parametric models or implicit parametric models trained on the digital wardrobes but failed to recover the garments with novel garment styles from the image. To generate novel garment styles from the input images, Deep Fashion3D~\cite{df3d} proposed to depict
garment styles with the feature lines predicted from the input image. Nevertheless, it fails to generate garment styles well aligned with the input images due to the inaccuracy of the boundary prediction. 
As for recovering surface details, MGN~\cite{bhatnagar2019multi} and SMPLicit~\cite{smplicit} can only produce smoothed garment mesh with limited surface details. BCNet~\cite{jiang2020bcnet} carves fine-grained details onto the garment template with an image-guided graph convolutional network though it fails to produce large-scale wrinkle deformations. Although Deep Fashion3D~\cite{df3d} can generate large-scale surface deformations based on Occupancy Network~\cite{mescheder2019occupancy}, the generated surface details may deviate from the input image as it only adopts global image features. Therefore, none of existing methods can recover garment styles and surface details aligning with the appearances from the input image.

The recent emergence of the pixel-aligned implicit~\cite{saito2019pifu,saito2020pifuhd,zheng2020pamir} framework has made it possible to reconstruct clothed humans with image-aligned appearances. On the other hand, it poses a question on how to exploit the power of the pixel-aligned framework to produce layered garment meshes that faithfully reflect the image appearances. 

To this end, we propose \sysname, a novel geometry inference framework that can produce high-fidelity layered garments by \textbf{[re]}gistering the \textbf{[e]}xplicit garment template meshes to the full-body implicit \textbf{[f]}ields predicted from single images. 
However, due to the diversity of the real-world garment geometry, 
it is non-trivial to establish correspondence between the garment template meshes and the clothing on an individual frame of the implicit clothed human.
To address this issue, we proposed novel methods to generate boundary fields and semantic fields to align the explicit garment template with the implicit clothed body. On top of the alignment, separated garment meshes with class-specific topology could be instantiated from the implicit fields with a dedicated designed optimization system. Experiments demonstrated that \sysnameb is capable of producing high-quality garment meshes from single images that could serve as off-the-shelf assets for various downstream applications, e.g., animation and simulation.

The main contribution of this work can be summarized as follows:

\begin{itemize}
    \item We proposed a novel geometry inference framework that reconstructs high-fidelity and topological-consistent garment meshes from single images by registering the explicit garment templates to an individual frame of the implicit clothed human body.
    \item We contribute to a novel learning-based method that predicts the implicit garment boundary fields with pixel-aligned features and curve-aligned features. The predicted garment boundaries can be well aligned to the appearances from the input image and the implicit clothed body.
    \item We conducted experiments on both synthetic datasets and in-the-wild datasets. The experiments demonstrated that our method could generate high-quality layered garments with accurate styles and expressive surface details.
\end{itemize}

\vspace{-4mm}
\section{Related Work}
\label{sec:related}
\begin{figure*}[!t]
\centering
\includegraphics[width=0.98\linewidth]{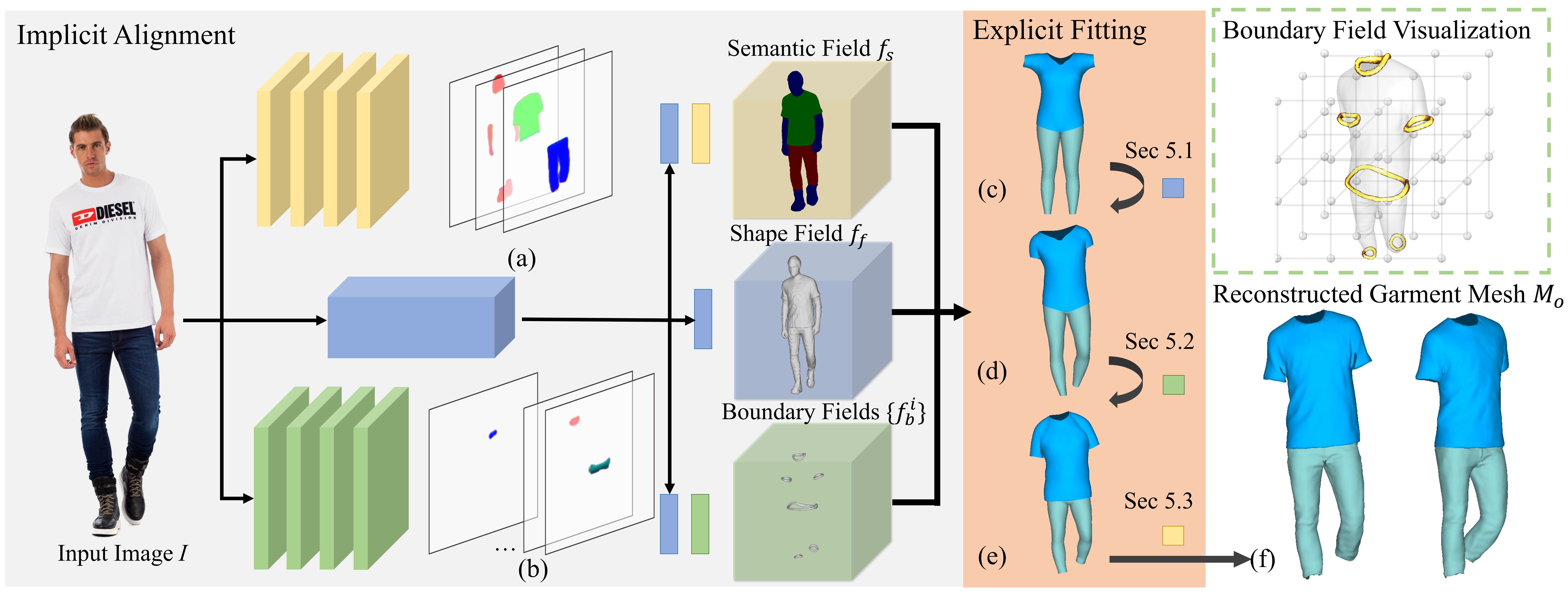}
\caption{ The pipeline of our proposed approach. (a) The semantic attention maps $\{H_{s}^{i}\}$ . (b) The boundary attention maps $\{H_{b}^{i}\}$ . (c) The explicit template mesh $M_{t}$. (d) The pose deformed template mesh $M_{p}$. (e) The boundary deformed template mesh $M_{l}$. (f) The output layered garments $M_{o}$.
}
\label{fig:realdataset}
\vspace{-4mm}
\end{figure*}

\subsection{Single View 3D Human Digitalization.}
Human digitalization from a single RGB image is inherently challenging due to the scarcity of information contained in the input regarding the diversity of the shape space. 
To make the ill-posed problem of single-view 3D human digitalization trackable,  SCAPE~\cite{anguelov2005scape} and SMPL~\cite{loper2015smpl} are proposed, which provide strong priors for later human-centric digitization tasks.
By simplifying the problem to low-dimensional body parameter estimation, ~\cite{anguelov2005scape,loper2015smpl,hmrKanazawa17,hasler2009statistical,pons2015dyna,joo2018total,lassner2017unite,omran2018neural,rhodin2016general,spec,pare} achieved human body and pose estimation from a single image.
However, these works based on parameterized models ~\cite{anguelov2005scape,loper2015smpl,hmrKanazawa17,hasler2009statistical,pons2015dyna,joo2018total,lassner2017unite,omran2018neural,rhodin2016general} are restricted to naked human body reconstruction. Since the garments and surface details are not modeled, the generated shapes are not suitable for visualization applications.

Thanks to the recent rising of 3D deep learning, many works on single-view human digitalization have been proposed to create high-quality 3D clothed human models~\cite{natsume2019siclope,varol2018bodynet,saito2019pifu,alldieck2019tex2shape,lazova3dv2019,alldieck19cvpr,pumarola20193dpeople,chen2013deformable,tang2019neural}. The works can be roughly divided into two streams: parametric methods and non-parametric methods. Parametric methods ~\cite{alldieck2019tex2shape, Zheng2019DeepHuman,tan2020self,monoclothcap,phygar} explicitly model the clothed human as the body parameters and the offset to the naked 3D parametric human body models. Though it can generate plausible results even from a single in-the-wild image, it fails to generate loose garments that are not close to the body.

Contrary to parametric-based methods, non-parametric models do not explicitly lean on the parametric human body and could reconstruct the clothed human body with arbitrary topologies. Siclope~\cite{natsume2019siclope} reconstructed clothed human from multi-view silhouette predicted from a single front-view image. DeepHuman~\cite{Zheng2019DeepHuman} achieved single image human reconstruction with an image-guided volume-to-volume translation network. Although both methods could generate human shapes with detailed garments of arbitrary topology, the details generated are relatively coarse or can not faithfully reproduce the input portrait’s appearances. Saito et al.~\cite{saito2019pifu,saito2020pifuhd,li2020monocular} addressed this issue through pixel-aligned implicit function and achieved high-fidelity reconstruction where the geometry generated can be pixel-wise aligned to the input images. Under the pixel-aligned framework, later works attacked the robustness of the model~\cite{zheng2020pamir,xiu2022icon}, or animating human encoded in the implicit space~\cite{arch,archplusplus}. However, the above methods fails to provide clothing mesh separated from the human body.

\subsection{Single View 3D Garment Reconstruction.} 
Compared with clothed 3D bodies, layered reconstruction of body and clothing provides easy-to-use assets for downstream tasks like animation and content creation. However, as garments are shapes with highly diversified topologies and complicated high-frequency surface details, separating garments from the human scan often involves laborious manual efforts. DeepWrinkles~\cite{lahner2018deepwrinkles} and Jin et al.~\cite{jin2018pixel} proposed synthesizing high-quality clothing wrinkle deformations as displacement maps or normal maps on the UV space of a fixed garment template, but they are not capable of handling large surface deformations and only support limited garment types.

Closer to our work are Multi-Garment Net(MGN)~\cite{bhatnagar2019multi}, SMPLicit~\cite{smplicit}, BCNet~\cite{jiang2020bcnet}, and Deep Fashion3D~\cite{jiang2020bcnet}. MGN pioneers by learning a per-category parametric model from a large-scale digital wardrobe. Layered garments could be inferred with MGN with a few images as input. SMPLicit~\cite{smplicit} introduced a generative model that supports reconstructing layered garments from a single parsed portrait. However, neither MGN nor SMPLicit can generate high-frequency details from single input images. 

To generate layered garments with vivid surface details, Deep Fashion3D~\cite{df3d} adopted Occupancy Network~\cite{mescheder2019occupancy} to reconstruct high-frequency surface details from the input image. The surface details generated by Occupancy Network are then transferred to smoothed template mesh with nonrigid-ICP. However, as Occupancy Network lean solely on global image features to produce surface details, it could not loyally recover the appearances from the input portrait. BCNet~\cite{jiang2020bcnet} firstly generates a coarse template mesh with PCA and then emboss surface details with an image-guided graph attention network but fails to produce large-scale wrinkle deformations. 

\vspace{-4mm}

\subsection{3D Shape Registration.}
3D Shape registration is a fundamental problem that has been extensively studied in the last decades, targeting setting up correspondences between predefined templates and novel observations. 
Previous works have tackled registering template mesh to explicit shapes(point clouds or meshes)~\cite{chui2003new} or registering implicit templates to implicit shapes~\cite{deng2021deformed,zheng2021deep}. Closer to our work, Clothcap~\cite{ponsmollSIGGRAPH17clothcap}, MetaAvatar~\cite{MetaAvatar:NeurIPS:2021}, SCALE~\cite{Ma:CVPR:scale}, and SCANimate~\cite{scanimate} could register garment templates or human body template to the clothed human scans, though they could only work on scan sequences. POP~\cite{POP:ICCV:2021} supports registering the animatable articulated dense point cloud to a single clothed human scan. LoopReg~\cite{bhatnagar2020loopreg} could be adopt for registering the parametric SMPL body model to a clothed human scan bridged by a space-diffused SMPL. MGN~\cite{bhatnagar2019multi} and Sizer~\cite{tiwari20sizer} register the garment templates to human scans though they can only handle tight clothing. Deep Fashion3D~\cite{df3d} registered a boundary-deformed garment template to the reconstructed garment mesh with non-rigid ICP. However, none of the methods mentioned above could register garment template meshes to whole-body implicit fields predicted from a single in-the-wild image. 

\section{Overview}
\label{sec:overview}
As illustrated in Figure 2., given a single in-the-wild image $I$, the goal of \sysnameb is to generate high-fidelity garment mesh $M_{o}$ with class-specific triangulation by registering the explicit garment template mesh $M_{t}$ to the predicted whole-body implicit field $f_{f}$. To this end, we decompose the whole registration process into two stages:~\textbf{aligning explicit to implicit}(Section~\ref{sec:inference}) and \textbf{fitting explicit to implicit} (Section~\ref{sec:fitting}). 
In the first stage, we will align explicit garment template mesh $M_{t}$ to the implicit target $f_{f}$ with implicit boundary fields $\{f_{b}^{i}\}$ and the implicit semantic field $f_{s}$ predicted from the input image $I$.
On top of the alignment, in the second stage, we will deform the explicit garment template $M_{t}$ to fit the implicit target $f_{f}$ with a dedicated designed optimization system.

\section{Aligning Explicit to Implicit}
\label{sec:inference}
Setting up accurate correspondence between the template and the target is the key to achieve a successful registration. In the following section, we will carefully introduce how to align the garment template mesh to the implicit clothed human body from the following aspects: the definition of explicit templates(Section~\ref{sec:extemp}), the generation of the implicit targets(Section~\ref{sec:imtar}), and the alignment between the explicit mesh and the implicit fields(Section~\ref{sec:boundinf}).

\subsection{Explicit template}
\label{sec:extemp}
We designed class-specific garment template meshes $M_t$ on top of the SMPL\cite{loper2015smpl} body following the previous works~\cite{bhatnagar2019multi,df3d,jiang2020bcnet}.
The designed garment template meshes $M_t$ covers 12 common clothes categories, including long/short/no sleeve uppers, long/short/no sleeve dresses, long/short/no sleeve open coats, long/short pants, and skirts. Notably, we define the outermost curves of each garment template $M_t$ as the garment boundaries$\{L_{t}^{i}\}$. It is worth mentioning that for garment templates that belong to the open coats categories, the necklines, the center-front lines, and hemlines are treated as different boundaries though they belong to the same curve. Please see the appendix for more details about the explicit garment template.
\subsection{Implicit Target}
\label{sec:imtar}
We adopt the pixel-aligned implicit framework to generate the implicit target $f_{f}$ (i.e., the implicit clothed human), which is superior to its counterparts by producing results that well match the input image.

\noindent \textbf{Pixel-aligned implicit framework} The pixel-aligned implicit framework~\cite{saito2019pifu,saito2020pifuhd} is built upon the implicit shape representation, where a 3D shape can be represented as the occupancy status within a bounded volume. Conditioned on an input image $I$, the pixel-aligned implicit function could predict the occupancy status of the queried coordinate $X\in R^3$ with:
\begin{equation}
   f(X,I) = g(X,\phi_{local}(I,\pi(X)))  
\end{equation}
where $\pi(X)\in(X_x,X_y)$ indicates the projected 2D position on image space. $\phi_{local}(I,\pi(X))$ denotes the image features fetched from the projected position.

\noindent \textbf{Implicit Target Generation.}
Inspired by the PIFu~\cite{saito2019pifu} and PIFuHD~\cite{saito2020pifuhd}, the learning process of fine-grained information, e.g., surface details and color, would be more trackable if conditioned on a coarse shape information descriptor. Therefore, we firstly defined the coarse shape field $f_{c}$ similar to the coarse branch in PIFuHD:
\begin{equation}
\begin{aligned}
   f_{c}(X,I) = &  g_{c}(X,\phi_{c}(I,\pi(X))) 
\end{aligned}
\end{equation}
where $\phi_{c}$ indicates the image features extracted from the down-sampled input.

To emboss the coarse shape field$f_{c}$ with fine-grained details, a fine shape field adopted as the implicit target $f_{f}$, is built on top of the coarse shape module. It takes the coarse shape embedding $\Omega_{c}(X)$ and fine-level image features $\phi_{f}$ to predict the occupancy status for the fine-grained shape:
\begin{equation}
   f_{f}(X,I^{\prime}) = g_{f}({\Omega_{c}(X)},\phi_{f}(I^{\prime},\pi^{\prime}(X)))  
\end{equation}
where $I^{\prime}$ denotes the cropped input image at the original resolution and $\pi^{\prime}(X)$ denotes the projected position of the sample points on the cropped image.

\subsection{Boundary Alignment}
\label{sec:boundinf}
We propose to set up boundary correspondence between the explicit template mesh $M_t$ and the implicit target $f_{f}$ for registration, as the boundaries possess the most prominent geometrical features of the garment shape. 
To obtain the garment boundaries on a 3D clothed human, one could turn to scan surface parsing or image-guided curve regression. However, as illustrated in Section.~\ref{sec:experiments}, the garment boundaries generated with surface parsing would be heavily corrupted due to the occlusion of the human body and other accessories. Although the image-guided curve regression can always deliver complete boundary curves, it fails to produce accurate boundary which aligns with the implicit target $f_{f}$. To this end, we propose a novel method that predicts a set of garment boundary fields  $\{f_{b}^{i}\}\in(-1,1)$ from the input image $I$, each representing a type of garment boundaries, e.g., collars, cuffs, hemlines.

\noindent \textbf{Garment boundary fields.} The garment boundaries are thin 3D spatial curves that are inherently hard to be captured by the implicit functions. Hence, instead of modeling each boundary curve directly with the implicit function, as illustrated in Figure.\ref{fig:realdataset}, we propose to model each garment boundary as an implicit cylinder with a signed distance field:
\begin{equation}
    f_{b}^{i}(X) = d(X - l_{b}^{i}) - \epsilon_{b}
\end{equation}
where the $d(X - l_{b}^{i})$ denotes the distance from the query point $X\in R^3$ to $i^{th}$ garment boundary. $\epsilon_{b}$ indicates the radius of the boundary cylinder that is set to $1e^{-3}$ empirically. 

\noindent \textbf{Vanilla approach.} We design a vanilla approach to predict the garment boundary fields $\{f_{b}^{i}\}$ from the input image $I$. To make sure that the predicted boundary fields $\{f_{b}^{i}\}$ are aligned with the target shape field $f_{f}$, we jointly trained the garment boundary fields $\{f_{b}^{i}\}$ and the target shape fields $f_{f}$ conditioned on the same coarse shape embedding $\Omega_{c}(X)$:

\begin{equation}
   f_{b}^{i}(X,I) = g_{b}^{vanilla}({\Omega_{c}(X)})  
\end{equation}

Although the vanilla approach can produce garment boundary fields $\{f_{b}^{i}\}$ that align with the target shape field $f_{f}$, it may not reflect the boundary appearances of the input image due to the lack of guidance received from the image space. 

\noindent \textbf{Curve-aligned boundary generation.} We thus propose a curve-aligned boundary generation module to generate garment boundary fields $\{f_{b}^{i}\}$ that accurately reflect the boundaries' appearances from the input image $I$. Compared with the vanilla approach which solely bases on pixel-aligned coarse shape feature $\Omega_{c}(X)$, our proposed curve-aligned boundary generation module may receive extra guidance, i.e., curve-aligned features, from the image space. 

To produce curve-aligned features for boundary field generation, we designed garment boundary attention maps which depict the likelihood of each garment boundary on image space. The garment boundary attention maps $\{H_{b}^{i}\}$ are generated from the input image $I$ with HigherHRNet~\cite{higher_hr} and could receive supervision from the ground-truth boundary heatmap. Conditioned on the curve-aligned features produced by boundary attention maps, the garment boundary fields can be generated with:
\begin{equation}
   f_{b}^{i}(X,I) = g_{b}({\Omega_{c}(X)},\phi_{h}(I,\pi(X)))  
\end{equation}
where $\phi_{h}(I,\pi(X))$ denotes the curve-aligned features sampled from the boundary attention maps$\{H_{b}^{i}\}$.

\subsection{Semantic Alignment.}
Apart from the boundary correspondence, semantic correspondence between the explicit template $M_{t}$ and the implicit target $f_{f}$ are required to mute the influences of non-relevant regions on the implicit target$f_{f}$. To this end, we designed semantic implicit fields $\{f_{s}^{i}(X,I)\}$, which denotes the occupancy likelihood for each kind of clothing(i.e., upper body clothing and lower body clothing) in 3D space. Notably, similar to the generation of garment boundary fields$\{f_{b}^{i}\}$, semantic attention maps $\{H_{s}^{i}\}$  predicted from the input images $I$ are adopted for additional 2D guidance:
\begin{equation}
   f_{s}^{i}(X,I) = g_{s}({\Omega_{c}(X)},\phi_{s}(I,\pi(X)))
\end{equation}

\noindent where $\phi_{s}(I,\pi(X))$ denotes the semantic attention map features fetched from the projected position $\pi(X)$. The semantic label for each 3D query point $X$ can be predicted by aggregating the implicit semantic fields of the possible labels:
\begin{equation}
   f_{s}(X,I) = \mathop{\arg\max}_{i}(f_{s}^{i}(X,I))
\end{equation}

\section{Explicit Fitting}
\label{sec:fitting}

In the previous section, we have bridged the gap between the explicit garment template $M_{t}$ and the implicit target(i.e., clothed human) with boundary correspondence $\{f_{b}^{i}\}$ and semantic correspondence $\{f_{s}^{i}\}$ predicted from the input image $I$. On top of the established correspondences, we proposed an explicit fitting pipeline, which progressively deforms the garment template mesh $M_{t}$ to be aligned with the implicit target $f_{f}$. The proposed explicit fitting pipeline consists of four phases, namely, template initialization(Section~\ref{sec:bodyinit}), boundary fitting(Section~\ref{sec:bounarydeform}), template fitting(Section~\ref{sec:bounarydeform}) and post processing(Section~\ref{sec:postprocessing}).

\subsection{Template Initialization} 
\label{sec:bodyinit}
As the explicit garment template mesh $M_{t}$ is built on top of the SMPL parametric human body~\cite{loper2015smpl}, accurate body pose and shape estimation may benefit the registration process by setting up a good initialization, i.e., the pose deformed garment template mesh $M_{p}$. However, estimating accurate 3D pose from a single in-the-wild image is inherently challenging due to the depth ambiguity, unknown camera parameters, and the scarcity of annotated in-the-wild datasets. In contrast, the state-of-the-art 2D pose estimation has reached relatively high accuracy on in-the-wild images. To this end, we propose to optimize the SMPL body parameters $SMPL(\theta,\beta)$ to be aligned with the implicit shape field under the additional guidance of selected 2D joints $J_{gt}$ predicted by off-the-shelf singe image pose estimator~\cite{cao2018openpose}:
\begin{equation}
\begin{aligned}
   V_{pred},J_{pred}  = & SMPL(\theta,\beta) \\
   {\mathcal L}_{body} = & MSE(J_{pred}^{\prime},J_{gt}) + \eta_{reg}Reg(\theta) \\ 
   & + \eta_{shape}CD(V_{lres},V_{pred})
\end{aligned}
\end{equation}
where $Reg$ denotes the pose regularization function adopted to reduce undesired poses and $V_{lres}$ indicates the low-resolution mesh vertices extracted from the coarse field $f_{c}$.

\subsection{Boundary Fitting} 
\label{sec:bounarydeform}

In Section~\ref{sec:boundinf}, we have established the correspondence between the  boundaries$\{l_{t}^{i}\}$ of the template mesh $M_{t}$ and the garment boundaries of the implicit target $f_{f}$ with the boundary fields $\{f_{b}^{i}\}$. Based on the boundary correspondence, we may deform the boundaries $\{l_{p}^{i}\}$ of the pose deformed template mesh $M_{p}$ to be aligned with the garment boundaries of the implicit target $f_{f}$:
\begin{equation}
   {\mathcal L}_{b} = f_{b}^{i}(l_{p}^i) + \eta_{ea}Avg(e_{b}^{i}) + \eta_{ed}Var(e_{b}^{i})
\end{equation}
where $e_{b}^{i}$ denotes the boundary edge lengths of the garment template mesh $M_{p}$. The optimized garment boundaries $\{l_{a}^{i}\}$ are set as the hard constraints for Bi-Harmonic deformation~\cite{jacobson2014bounded}. So far, a plain garment template mesh $M_{l}$ is produced with the garment boundaries $\{l_{a}^{i}\}$ aligned with the garment boundaries of the implicit target $f_{f}$.

\subsection{Shape Fitting}
\label{sec:shapefitting}
By the end of the boundary fitting stage, a plain garment template mesh $M_{l}$ is generated whose boundaries are aligned to the garment boundaries of the implicit target $f_{f}$. To emboss the boundary-aligned plain template mesh $M_{l}$ with fine-grained details, we compiled an optimization system with the following design goals: Firstly, the resulting garment $M_{o}$ mesh should stick tightly with the corresponded part on the implicit target $f_{f}$. Secondly, the boundaries of the resulting mesh $\{l_{o}^{i}\}$ should remain aligned with the garment boundaries of the implicit target $f_{f}$. Thirdly, the resulting mesh $M_{o}$ should not penetrate with the predicted human body mesh $M_{smpl}$.

However, as the implicit target $f_{f}$ encodes the whole clothed human body, directly fitting the explicit template $M_{p}$ to the implicit target $f_{f}$ may be affected by non-relevant regions, i.e., hairs, skins, or other clothes. To address this issue, we propose an operation called activate area probing, which predicts the activated template vertices to be deformed, and the approximated  distances $D_{act}$ between the activated vertices to the corresponding areas on the implicit target $f_{f}$.

\begin{figure}[!t]
\centering
\includegraphics[width=1.0\linewidth]{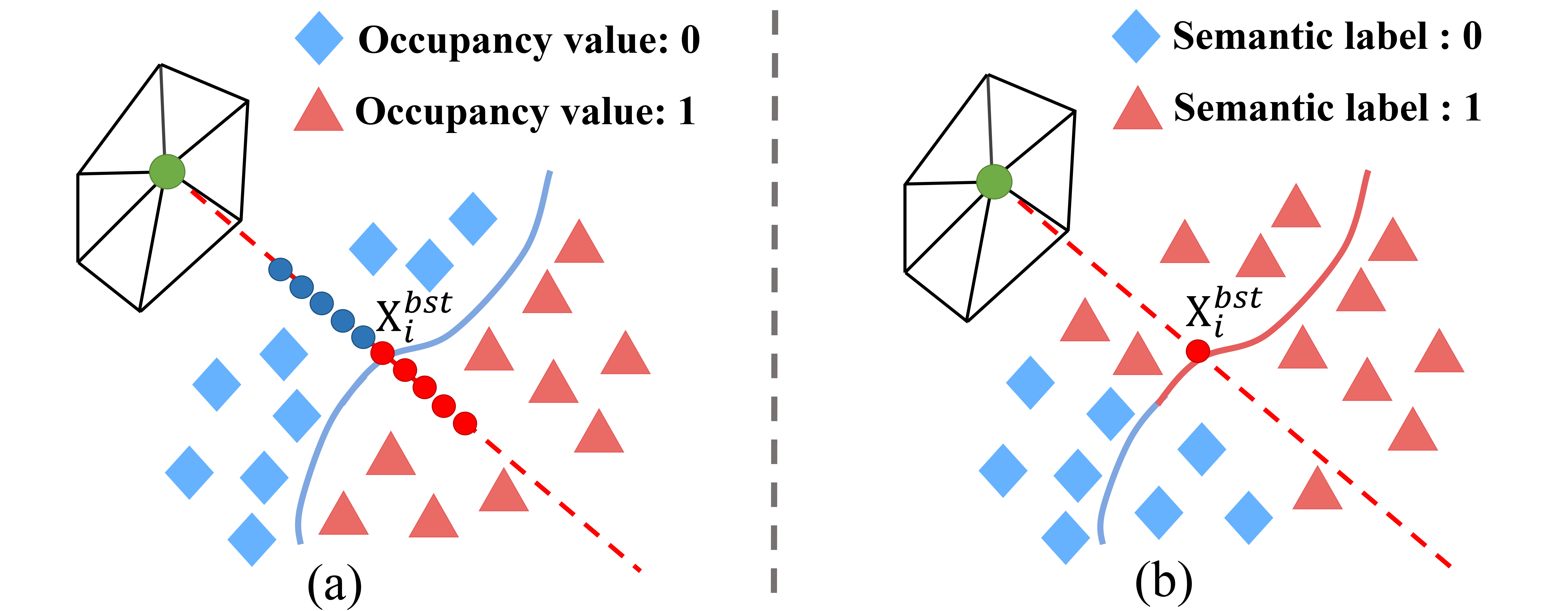}
\vspace{-3mm}
\caption{A illustration of our proposed active area probing scheme, our proposed active area probing scheme could aggregate the implicit shape information and implicit semantic information to guide the deformation of the garment template mesh.}
\vspace{-4mm}
\label{fig:tracing}
\end{figure}
\noindent \textbf{Activate Area Probing.} Given a vertex $X_{i}$ on the explicit template $M_{l}$, we cast rays in both directions along the vertex normal to sample $k$ points on each direction. 
For each vertex $X_{i}$, 2k + 1 points $\{X_{i}^{0},X_{i}^{1},...,X_{i}^{2k-1},X_{i} \}$ in total are fed to both implicit target field $f_{f}$ and implicit semantic fields $\{f_{s}^{i}\}$ in batches. The approximate distance can be calculated as the distance from the template vertex $X_{i}$ to the closet sample point $X_{i}^{bst}$ that penetrates the surface threshold $\epsilon$ (i.e. $0.5$). The activation status $B_{i} \in \{0,1\}$ for the vertex $X_{i}$ is set to active only when the cast rays reach the iso-surface and the semantic label at the penetration point $f_{s}(X_{i}^{bst})$ is consistent with the current template. Finally, the approximated distance from the activate areas on of the explicit template mesh $M_{t}$ to the implicit target $f_{f}$ can be calculated with:
\begin{equation}
   D_{act}(M_{o}) = Avg(B_{i} \dot MSE(X_{i}, X_{i}^{bst}))/Sum(B_{i})
\end{equation}

With the proposed activate region loss $D_{act}(M_{o})$, we can update the explicit template mesh $M_{o}$ to fit corresponding areas on the implicit target $f_{f}$ with the following loss function:
\begin{equation}
\begin{aligned}
{\mathcal L}_{o} = D_{act}(M_{o}) - \eta_{pen}TSDF(M_{smpl})(M_{o}) \\+ \eta_{b}{\mathcal L}_{b} + \eta_{lap}{\mathcal L}_{lap}
\end{aligned}
\end{equation}
where $TSDF(M_{smpl})$ indicates the truncated signed distance function of the posed human body mesh $M_{p}$ adopted for penalizing the garment-body penetration, and ${\mathcal L}_{lap}$ denotes the laplacian of the deformed template mesh. By the end of the explicit fitting stage, we will obtain a high-fidelity garment mesh $M_{o}$ well align with the input image $I$.

\subsection{Post Processing}
\label{sec:postprocessing}
While the reconstructed garment meshes $M_{o}$ could well recover the garment styles and surface details from an in-the-wild input image, like most existing image-based reconstruction methods, it may fail to reconstruct folded structures like the collars. 
Therefore, we manually created a collar warehouse containing various real-world collars built upon the garment template and trained a light-weight image classification network to choose the collar type with the closest appearance to the image.
Thanks to the topology-consistent nature of our generated garment mesh, the collar can be attached to the garment template through vertex correspondence. The collar's geometry is further tuned with Bi-Harmonic deformation to be collocated with the reconstructed garment mesh.

\section{Experiment Results}
\label{sec:experiments}
\subsection{Implementation Details}
\noindent\textbf{Data Preparation} 
We adopt RenderPeople~\cite{RenderPeople} data to train our proposed model, which contains 400 photo-realistic 3D clothed humans with high-resolution textures and surface semantic parsing. 
We split the whole dataset into a training set of 360 subjects and a testing set of 40 subjects. 
All of the textured scans are rendered following the settings in PIFuHD~\cite{saito2020pifuhd}.
It is worth mentioning that although the semantic parsing provided by RenderPeople could help to identify the garment boundaries automatically, they may be heavily corrupted due to the occlusion of the human body and accessories. Therefore, we hired professional artists to annotate the garment boundaries on the scan surfaces. 
More importantly, the artists may link the incomplete boundary segments into smoothed closed curves with their expertise in garments' shape. 

\noindent\textbf{Network Training}
The coarse shape, boundary, and semantic field generation modules are trained with the input image rescaled to $512\times 512$. The target shape field generation module is trained with random cropped images at the original resolution with window size as $512\times512$. We jointly train the coarse shape generation module, the boundary field generation module, and the semantic field generation module with a learning rate of $1\times10^{-4}$ for six epochs from scratch. The fine shape module is trained conditioned on a fixed coarse shape module with a learning rate of $1\times10^{-4}$. It takes roughly 72 hours to train all the modules mentioned above on two GTX 3090 GPUs. Please refer to the appendix for more implementation details about the network training and the explicit fitting.

\subsection{Ablation Studies}
In this section, we compile a set of ablation experiments to verify the algorithmic components’ effectiveness for our boundary field generation module. Please refer to the appendix for more details on the ablations for the explicit fitting stage.
 
\begin{figure}[!t]
\centering
\includegraphics[width=0.9\linewidth]{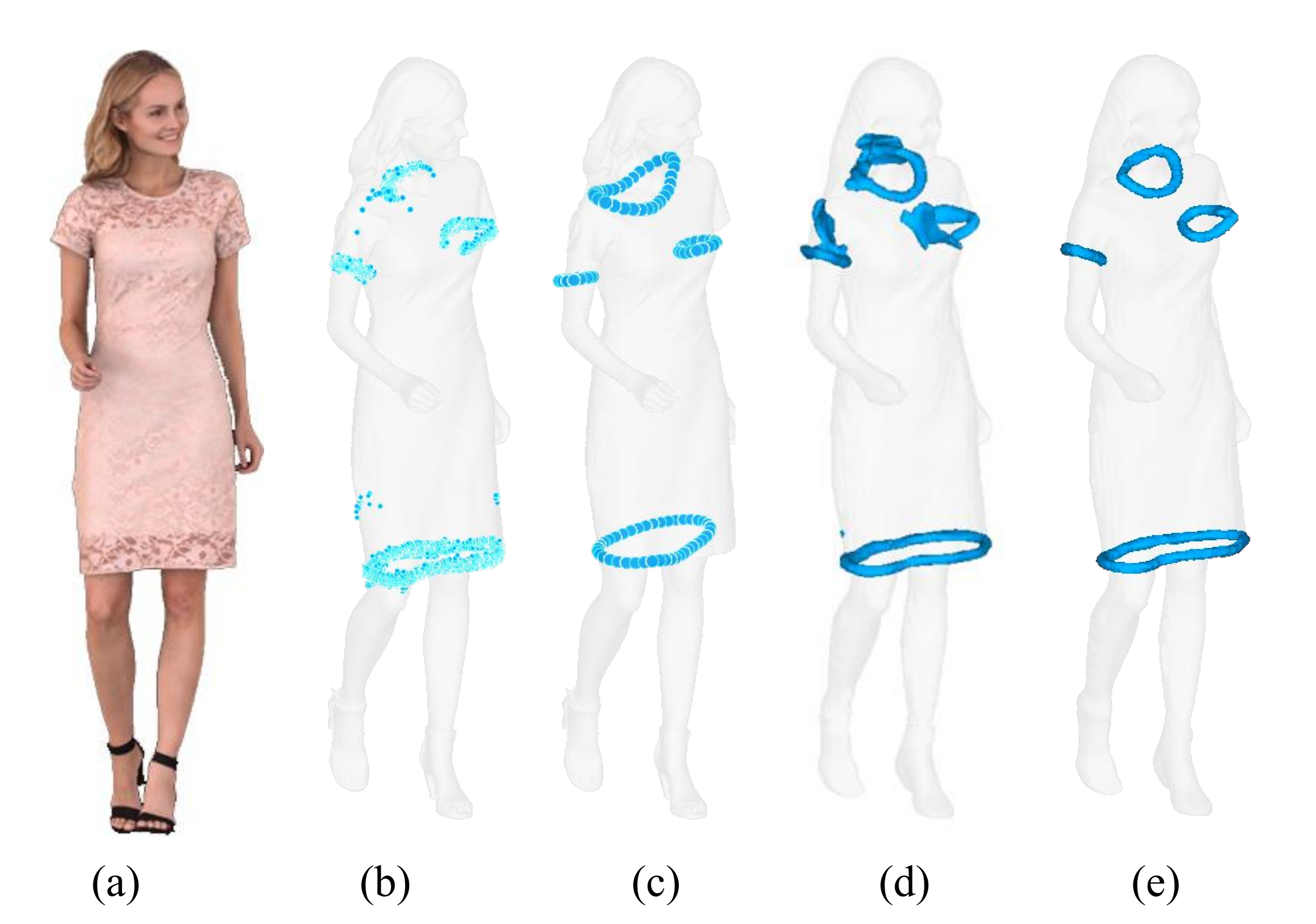}
\vspace{-3mm}
\caption{
Qualitative comparison of the garment boundaries produced under different ablation settings. The input image (a) is followed by the garments generated with (b) \textbf{PCT}, (2) \textbf{GCN}, (3) \textbf{w/o HM} and (4) \textbf{Ours}.
\vspace{-4mm}
}
\label{fig:qualablation}
\end{figure}

We orchestrated both quantitative and qualitative comparisons between our proposed model and the alternatives that take other candidate design choices: 1) Predict the garment boundaries by parsing the dense point cloud sampled on the explicit surface with Point Transformer~\cite{zhao2021pointtransformer}, termed as \textbf{PCT}. 2) Predict the garment boundaries by regressing the explicit curves with an image-guided graph convolution network, termed as \textbf{GCN}. 3) Predict the garment boundaries with pixel-aligned coarse shape features identical to the vanilla approach mentioned in Sec.\ref{sec:boundinf}, termed as \textbf{w/o  HM}. 4) The proposed full model, termed as \textbf{Ours}. Specifically, we extracted explicit mesh from the garment boundary fields generated with setting \textbf{w/o  HM} and \textbf{Ours} with Marching Cubes~\cite{lorensen1987marching} for later comparison. Table.\ref{tab:featlineacc} shows the quantitative comparisons between the design alternatives and the proposed one. As seen, the proposed approach exhibits the best accuracy among all the settings.

~\begin{table}[ht]
\small
    \centering
    \begin{tabular}{l|l|l|l|l}
    \toprule
     Methods & \textbf{PCT} & \textbf{GCN} & \textbf{w/o HM} & \textbf{Ours} \\
    \midrule
     CD($\times 10^{-3})$ & 6.5329 & 9.18467 & 6.3786 & \textbf{1.1073} \\
    \bottomrule
    \end{tabular}
    \vspace{-2mm}
    \caption{The quantitative comparison between the proposed model and the ablation alternatives.}
    \label{tab:featlineacc}
\end{table}
\vspace{-4mm}

Figure.\ref{fig:qualablation} shows the visualization results generated under different experiment settings. As clothed human bodies are highly diversified shapes with various surface details, \textbf{PCT} may produce corrupted garment boundaries and noisy parsing. Though \textbf{GCN} can produce complete curves, the boundary curves generated with \textbf{GCN} are largely deviates the clothing boundary.  Due to the lack of the guidance from the image space, \textbf{w/o HM} may generate garment boundaries with undesired shapes. \textbf{Ours} can produce clean garment boundaries that are well aligned with the boundary appearances from the images.

\begin{figure*}[t]
\centering
\includegraphics[width=0.95\linewidth]{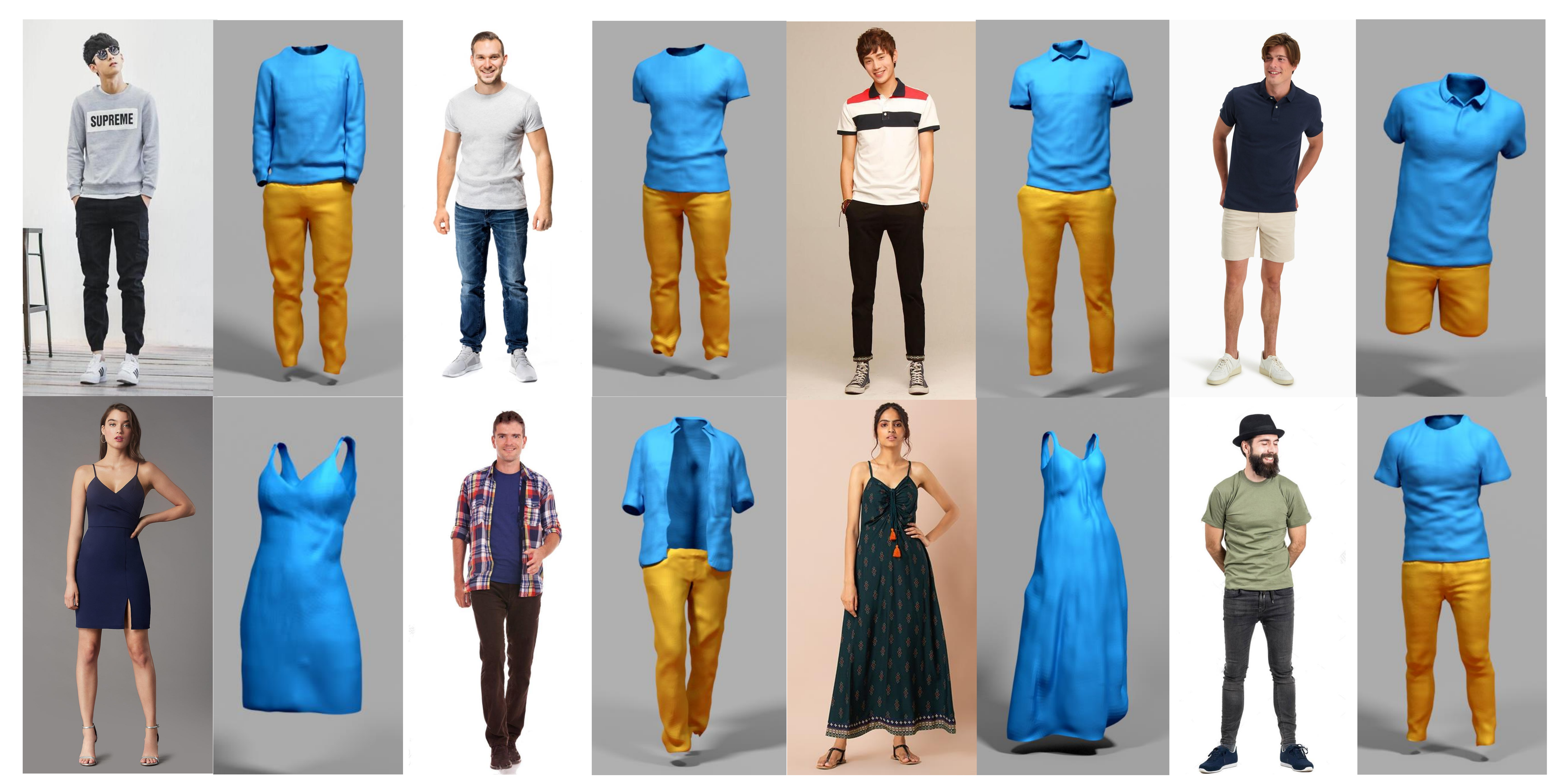}
\vspace{-3mm}
\caption{The results generated by our method on in-the-wild images. Each image is followed by the reconstructed layered garment mesh.
}
\vspace{-4mm}
\label{fig:gallary}
\end{figure*}

\subsection{Comparison Experiments} 
We compared our method with the state-of-the-art single image garment reconstruction methods of which the codes are publicly available, i.e., Multi-Garment Net~\cite{bhatnagar2019multi}, BCNet~\cite{jiang2020bcnet}, and SMPLicit~\cite{smplicit}, both quantitatively and qualitatively. 

\noindent \textbf{Quantitative Comparison} We test our method and the state-of-the-art methods with the rendered images from our synthetic  testing set. Notably, the garment meshes generated by different methods are aligned to the ground truth garment meshes with the underlying SMPL body. After aligning the results to the ground truth garment mesh, we compute the Chamfer Distance (CD) between the reconstructed mesh and ground truth for accuracy measurement. As illustrated in Table.\ref{tab:quantitive}, our method outperforms the comparison counterparts in reconstruction accuracy by a large margin.

\begin{table}[ht]
\small
    \centering
    \begin{tabular}{l|l|l|l|l}
    \toprule
     Methods & MGN & SMPLicit & BCNet & Ours \\
    \midrule
     CD($\times 10^{-3})$& 1.1424 &  1.3408 & 0.9725 & \textbf{0.5477} \\
    \bottomrule
    \end{tabular}
    \vspace{-2mm}
    \caption{The quantitive comparison between our model with the state of the art garment reconstruction methods. }
    \vspace{-3mm}
    \label{tab:quantitive}
\end{table}

\begin{figure}[!t]
\centering
\includegraphics[width=0.98\linewidth]{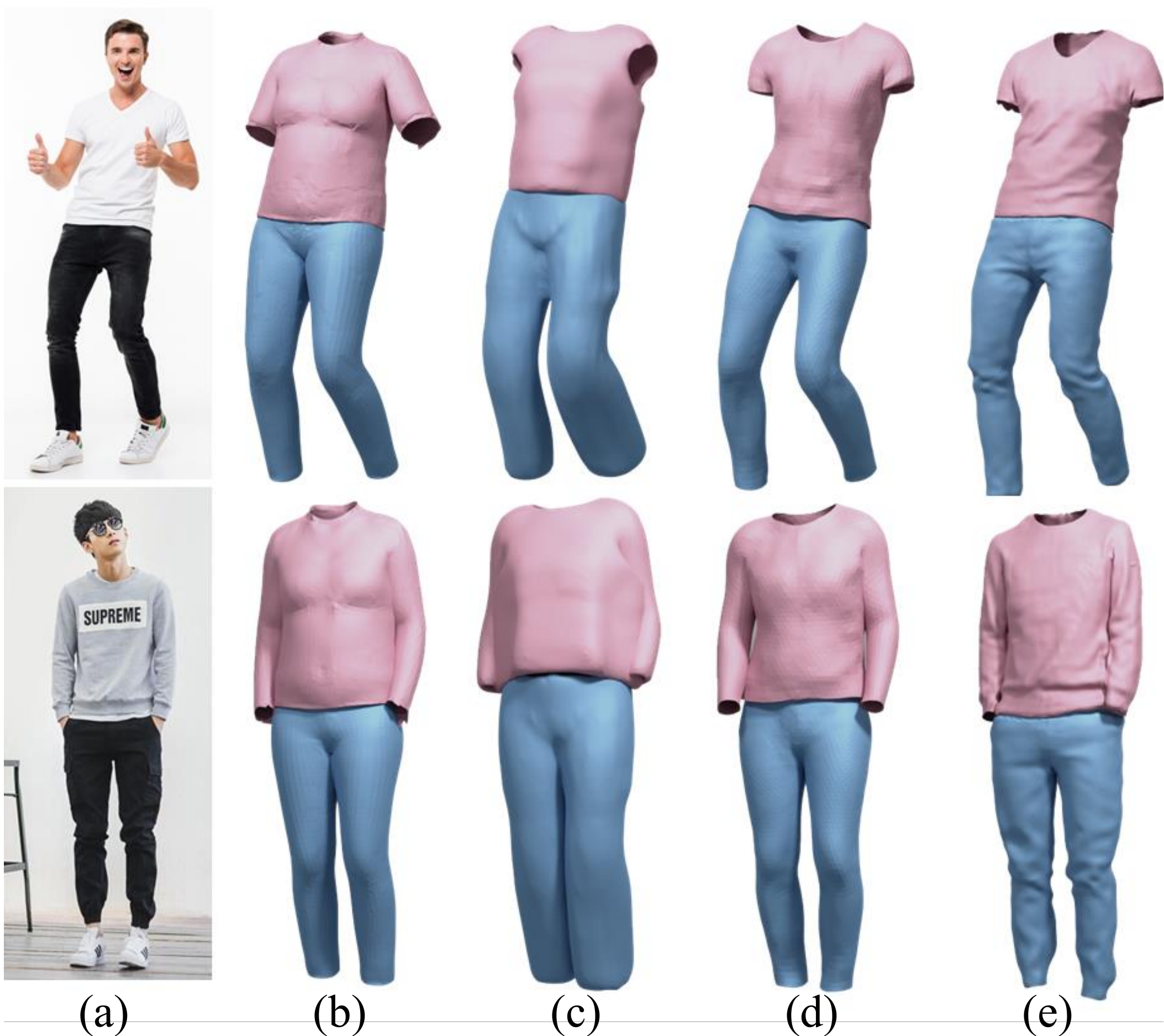}
\vspace{-2mm}
\caption{
 Qualitative comparison between ours and the state of the arts.
 For each row, the input image (a) is followed by the results generated by (b) Multi-Garment Net~\cite{bhatnagar2019multi}, (c) SMPLicit~\cite{smplicit}, (d) BCNet~\cite{jiang2020bcnet} and (e) our method.
}
\vspace{-4mm}
\label{fig:comparenb}
\end{figure}

\begin{figure}[!t]
\centering
\includegraphics[width=1.0\linewidth]{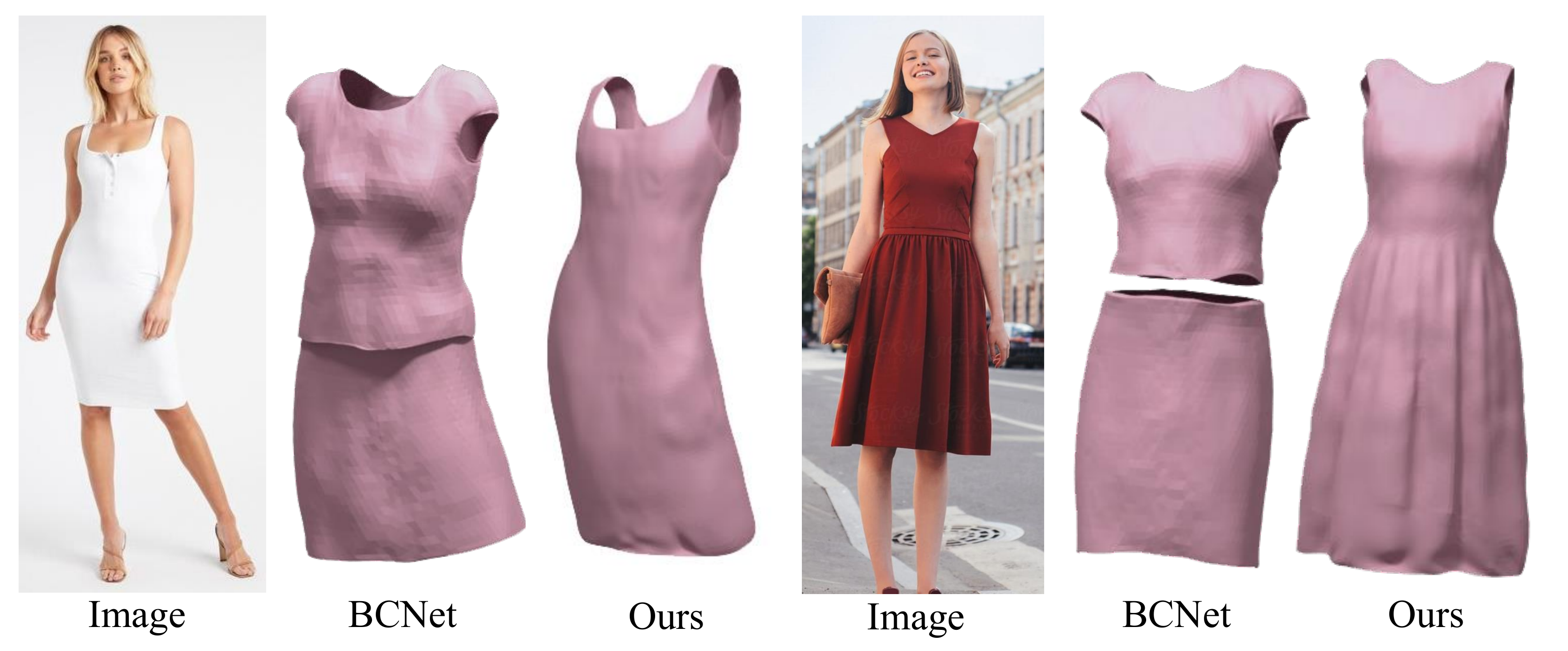}
\vspace{-2mm}
\caption{The qualitative comparison between our proposed method and BCNet~\cite{jiang2020bcnet} on dress reconstruction.}
\vspace{-4mm}
\label{fig:comparend}
\end{figure}

\noindent \textbf{Qualitative Comparison} 
Figure.\ref{fig:comparenb} provides qualitative comparisons on the results generated with in-the-wild images collected from the internet. Compared to the other methods, our method is superior in reconstructing accurate garment styles and reproducing the surface details faithfully. Figure.\ref{fig:comparend} demonstrates the qualitative comparison between our method and BCNet~\cite{jiang2020bcnet}. While BCNet\cite{jiang2020bcnet} fails to produce garments with the correct style, our proposed method could reconstruct the garment meshes with boundaries and surface details highly identical to the image input.

\subsection{Gallery on in-the-wild images}
Figure.\ref{fig:gallary} shows the results generated by our proposed method on in-the-wild images. The results demonstrate that our method could produce high-quality garments with fine grained details and correct garment styles.

\section{Conclusion and Limitation}
\label{sec:conclusion}
Layered garment reconstruction from a single in-the-wild image is inherently a challenging problem due to the highly diversified garment shapes and the high-frequency details. To this end, we proposed a novel pipeline which faithfully recovers high-quality garments from a single image by registering the explicit mesh to the implicit fields. A novel garment boundary field generation model is proposed to align the explicit template mesh to the implicit target. Based on the seamless alignment between the explicit and the implicit, the proposed pipeline may produce high quality garment meshes through fitting template mesh to the corresponding parts on the implicit target.

\noindent \textbf{Limitations} Although our methods may generate high quality garments from a single image. It only supports reconstructing clothing in common clothes categories. In the future, we will attack the problem of the generation of clothing with complex topology and multi-layered clothing.

\noindent \textbf{Acknowledgement} 
The work is supported by the Basic Research Project No.HZQB-KCZYZ-2021067 of Hetao Shenzhen-HK S\&T Cooperation Zone, National Key R\&D Program of China with grant No.2018YFB1800800 by Shenzhen Outstanding Talents Training Fund 202002, and by Guangdong Research Projects No.2017ZT07X152 and No.2019CX01X104. It is also supported by NSFC-62172348, 61902334 and Shenzhen General Project  (JCYJ20190814112007258). We thank the ITSO in CUHKSZ for their High-Performance Computing Services.

{\small
\bibliographystyle{ieee_fullname}
\bibliography{latex.bib}
}

\clearpage
\setcounter{figure}{0}
\setcounter{table}{0}
\appendix

In this appendix, we provide more results and details in the following aspects: (1) more implementation details regarding the generation of explicit template mesh, network training, and explicit fitting; (2) more evaluation on the explicit fitting stage; (3) more results reconstructed from the in-the-wild images.

\section{Explicit Template}
\label{sec:suppexplicit}
\begin{figure}[!t]
\centering
\includegraphics[width=0.9\linewidth]{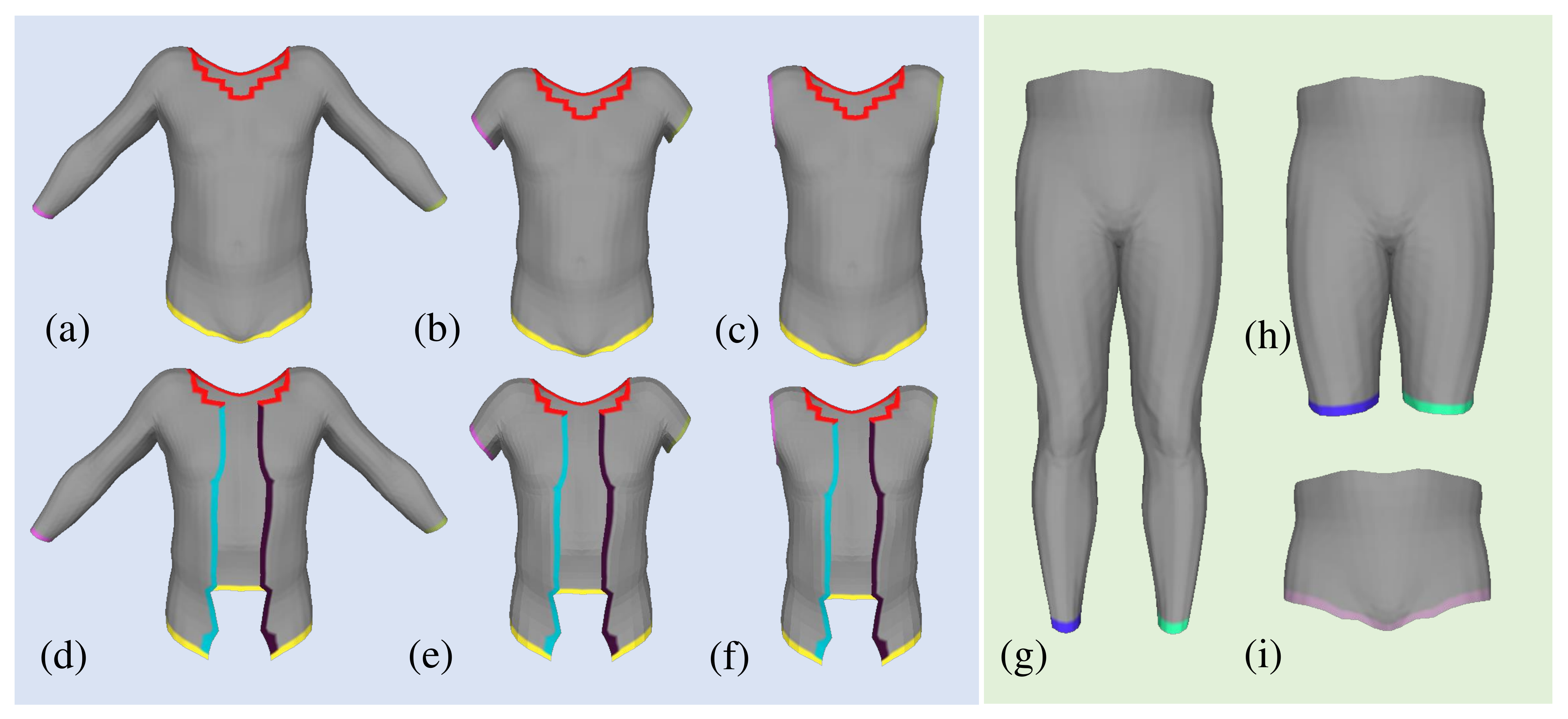}
\caption{
Garment templates supported by \sysname: (a) long-sleeve upper clothing and long-sleeve dress; (b) short-sleeve upper clothing and short-sleeve dress; (c) no-sleeve upper clothing and no-sleeve dress; (d) long-sleeve open coat; (e) short-sleeve open coat; (f) no-sleeve open coat; (g) long pants; (h) short pants; (e) skirt. Different kinds of garment boundaries are annotated with distinct colors.
}
\label{fig:supptemp}
\end{figure}

As is illustrated in Figure.\ref{fig:supptemp}, the explicit garment template meshes ${M_t}$ in \sysnameb covers twelve common clothes categories. On top of the garment template meshes $M_t$, we defined nine types of garment boundaries $\{L_{t}^{i}\}$. Each type of garment  boundary is corresponded to a boundary implicit boundary field $\{ f_{b}^{i}\}$.

\section{Network Training}
\label{sec:supptrain}
In the main paper, we have briefly introduced the generation of the implicit target shape field $f_{f}$, implicit semantic fields $\{f_{s}^{i}\}$ and the implicit boundary fields $\{f_{b}^{i}\}$. This section will describe the detailed training settings for the implicit field generation networks.

\subsection{Attention map generation.}
As mentioned in the main paper, generating implicit semantic fields $\{f_{s}^{i}\}$ and implicit boundary fields $\{f_{b}^{i}\}$ requires curve-aligned features $\phi_{h}(I,\pi(X))$ fetched from the predicted semantic and boundary attention maps. To this end, we generate ground truth boundary heat maps and semantic heat maps as the supervision to guide the network's training: Firstly, we project the points sampled on different semantic regions(or boundaries cylinders) to different image planes with a weak-perspective camera. Then, we generate Gaussian kernels centering at the projected positions with $\sigma$ set to 2. Finally, the semantic/boundary heat maps($\{H_{s}^{i}\}$ and $\{H_{b}^{i}\}$) can be obtained by fusing the Gaussian kernels with maximum operator on each image plane.

\subsection{Loss Functions.}
As mentioned in the main paper, we jointly train the generation module for coarse shape field $f_{c}$, the semantic fields $\{f_{s}^{i}\}$ and the boundary fields $\{f_{b}^{i}\}$ with coarse occupancy loss ${\mathcal L}_{cocc}$, semantic attention loss ${\mathcal L}_{hms}$, boundary attention loss ${\mathcal L}_{hmb}$, boundary field loss ${\mathcal L}_{b}$ and semantic field loss ${\mathcal L}_{s}$ :

\begin{equation}
\begin{aligned}
   {\mathcal L} = {\mathcal L}_{cocc} + {\mathcal L}_{hms} + {\mathcal L}_{hmb} + {\mathcal L}_{b}  + {\mathcal L}_{s}
\end{aligned}
\end{equation}
where the loss for each component is the mean squared error(MSE) between the predicted value and the ground truth. 
 
\section{Explicit Fitting}
\label{sec:suppfit}
In the main paper, we have explained the loss functions adopted for deforming the explicit template progressively to fit the implicit shape. On this top, we will provide further details on the explicit fitting regarding the hyper-parameter settings and the post processing.

\noindent \textbf{Template initialization.} With the purpose of setting up a good initialization for the later stages, we optimized the SMPL body parameters $SMPL(\theta,\beta)$ to be aligned with the implicit clothed body and the predicted 2D joints $J_{gt}$:
\begin{equation}
\begin{aligned}
   V_{pred},J_{pred}  = & SMPL(\theta,\beta) \\
   {\mathcal L}_{body} = & MSE(J_{pred}^{\prime},J_{gt}) + \eta_{reg}Reg(\theta) \\ 
   & + \eta_{shape}CD(V_{lres},V_{pred})
\end{aligned}
\end{equation}
where $\eta_{reg}$ is set to $1e^{-3}$ and $\eta_{shape}$ is set to $1.0$.

\noindent \textbf{Boundary Fitting.} To further fit the initialized template to align with the garment boundaries of the implicit target, we may deform the boundaries by minimizing the following loss function:
\begin{equation}
   {\mathcal L}_{b} = f_{b}^{i}(l_{p}^i) + \eta_{ea}Avg(e_{b}^{i}) + \eta_{ed}Var(e_{b}^{i})
\end{equation}
where $\eta_{ea}$ is set to $0.025$ and $\eta_{ed}$ is set to $2.5$.

\noindent \textbf{Shape Fitting} The shape fitting stage will further deform the boundary-aligned garment template to approach the implicit target guided by the following loss function:

\begin{equation}
\begin{aligned}
{\mathcal L}_{o} = D_{act}(M_{o}) - \eta_{pen}TSDF(M_{smpl})(M_{o}) \\+ \eta_{b}{\mathcal L}_{b} + \eta_{lap}{\mathcal L}_{lap}
\end{aligned}
\end{equation}

where $\eta_{pen}$, $\eta_{b}$ and $\eta_{lap}$ are set to $0.1$, $0.1$ and $100$ respectively.

\begin{figure}[!t]
\centering
\includegraphics[width=0.98\linewidth]{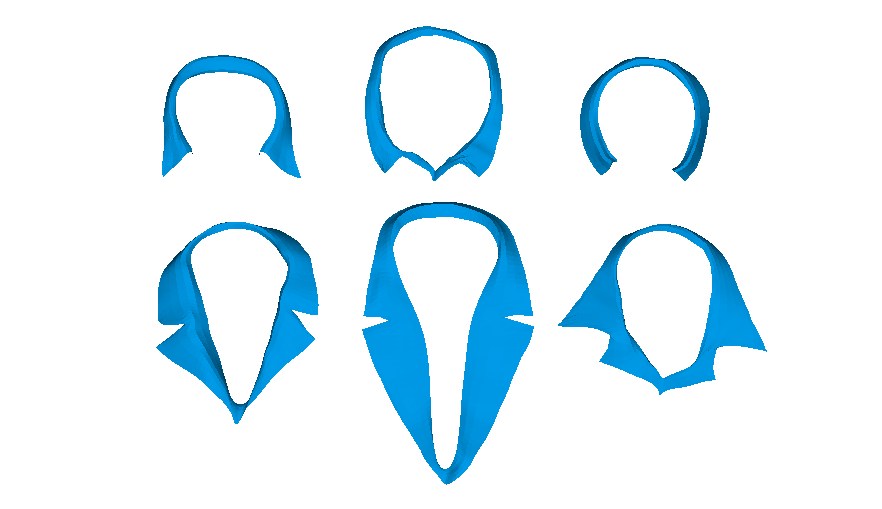}
\caption{
Selected collar templates from our collar warehouse.
}
\label{fig:suppcollar}
\vspace{-1em}
\end{figure}

\noindent \textbf{Post Processing} As mentioned in the main paper, our method could recover the garment styles and surface details from an in-the-wild input image though it may fail to generate folded structure, i.e., the collars. Therefore, as Figure.\ref{fig:suppcollar} illustrates, we firstly create a collar warehouse that covers ten common collars categories. A multi-layer perceptron is then adopted, which takes image features(for coarse shape field generation) sampled from the collar area to predict the type of the collar presents on the image. 

\section{Ablation Study}
\label{sec:suppablation}
In this section, we compile a set of ablation experiments to verify the effectiveness of each algorithmic component for our explicit fitting module. We provide qualitative comparisons between our proposed method and the alternatives that take other candidate settings: 1) Deform the garment template mesh to fit the implicit target without pose initialization, termed as \textbf{w/o Init}. 2) Deform the garment template mesh to fit the implicit target without boundary initialization, termed as \textbf{w/o Bound}. 3) Deform the garment template mesh to fit the implicit target without active area probing, termed as \textbf{w/o Probe}. 4) The proposed full model, termed as \textbf{Ours}. 

\begin{figure}[h]
\centering
\includegraphics[width=0.98\linewidth]{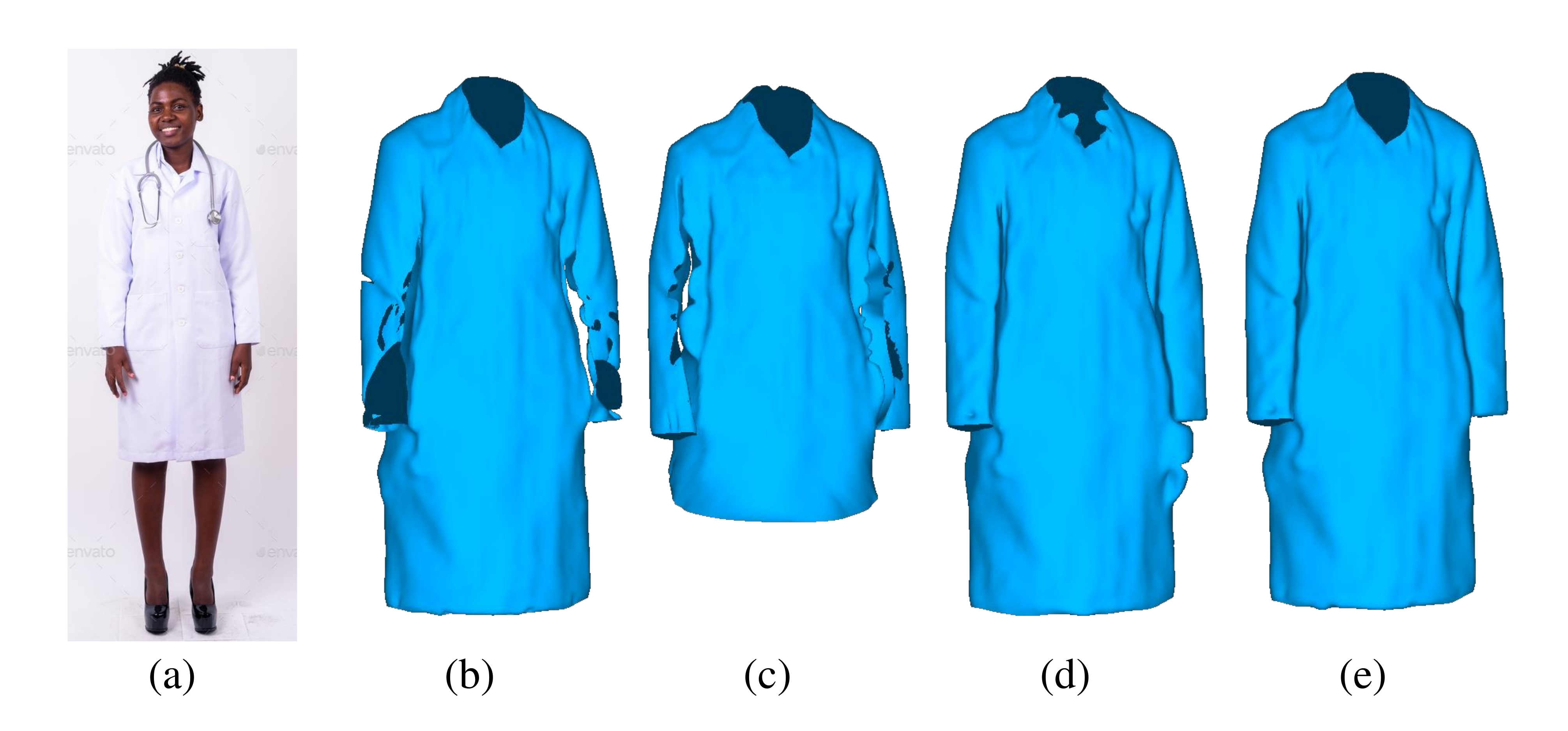}
\caption{
Qualitative comparison of the explicit garment meshes generated under different ablation settings. The input image (a) is followed by the garments generated with (b) \textbf{w/o Init}, (c) \textbf{w/o Bound}, (3) \textbf{w/o Bound} and (4) \textbf{Ours}.
}
\label{fig:suppablation}
\end{figure}
\vspace{-1em}
Figure.\ref{fig:suppablation} and Table.\ref{tab:ablacc} demonstrate the qualitative and quantitative comparison between the proposed model and the design alternatives. As the garments are diversified shapes with varying geometrical details, it is inherently hard to strike a balance between the reconstruction accuracy and surface smoothness without proper initialization(\textbf{w/o Init} and \textbf{w/o Bound}). Although the results generated with \textbf{w/o Probe} can well reflect the garment styles and most surface details from the image, the reconstructed surface would be corrupted by non-relevant regions (e.g. the hands for this case). In contrast, \textbf{Ours} can produce high-quality garment meshes with accurate styles and surfaces details highly identical to the input image.

\begin{table}[h]
\small
    \centering
    \begin{tabular}{l|l|l|l|l}
    \toprule
     Methods & \textbf{w/o Init} & \textbf{w/o Bound} & \textbf{w/o Probe} & \textbf{Ours} \\
    \midrule
     Dist($\times 10^{-3})$ & 3.52109 & 70.3211 & 3.56883 & \textbf{3.41651} \\
    \bottomrule
    \end{tabular}
    \vspace{-0.8em}
    \caption{Comparison on registration accuracy between the proposed method and the ablation alternatives.}
    \label{tab:ablacc}
\end{table}
\vspace{-4mm}

\section{More results on in-the-wild images}
\label{sec:suppresults}
\begin{figure*}[t]
\centering
\includegraphics[width=0.95\linewidth]{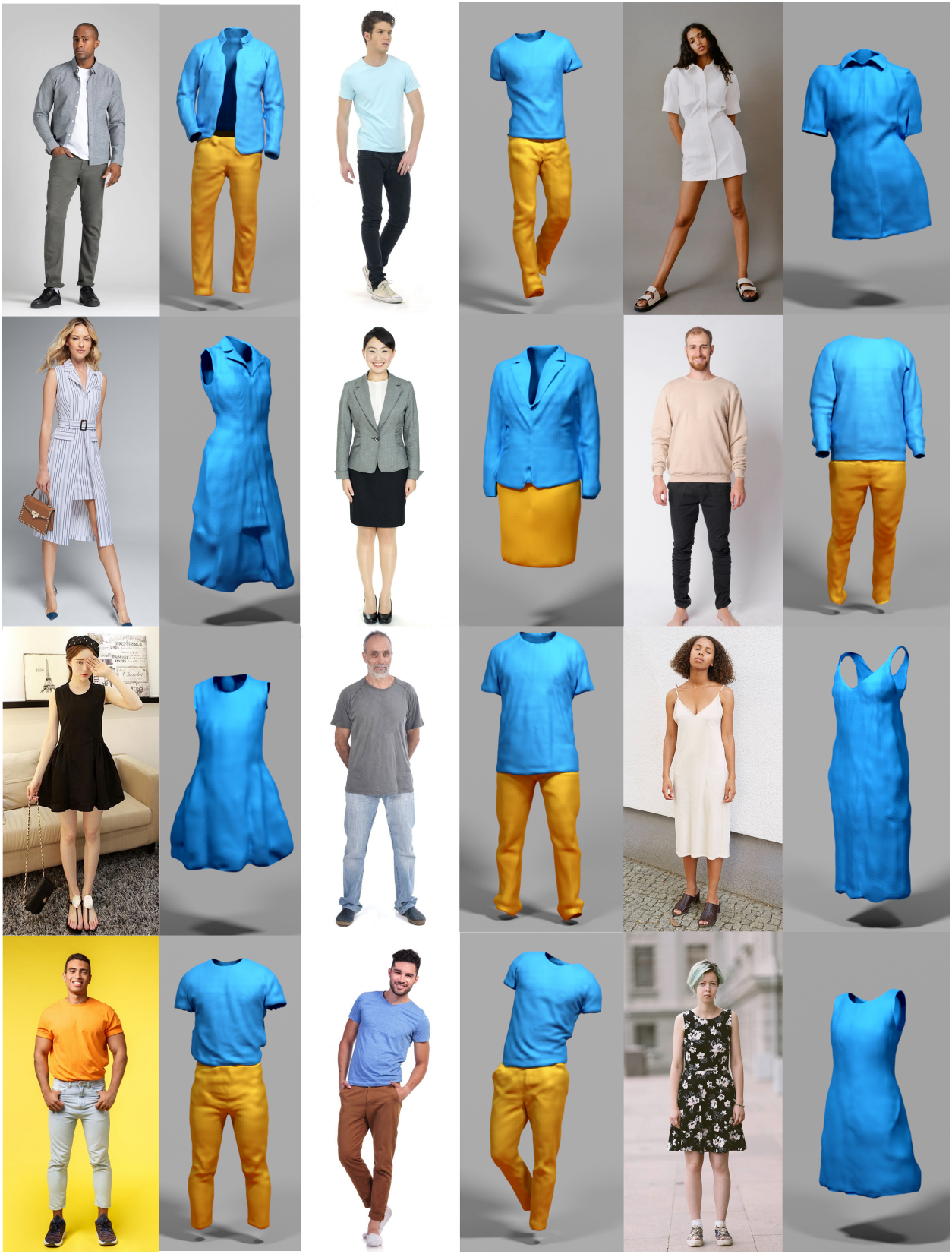}
\vspace{-1mm}
\caption{The results generated by our method on in-the-wild images. Each image is followed by the reconstructed layered garment mesh.
}
\vspace{-1mm}
\label{fig:suppgallary}
\end{figure*}

This section provides more results generated by our method on in-the-wild images from the internet. As is shown in Figure.\ref{fig:suppgallary}, given an in-the-wild image as input, our model could produce high-quality garments with fine-grained details and correct garment styles.

\end{document}